\documentclass[letterpaper, 10 pt, conference]{ieeeconf}  

\usepackage{cite}
\usepackage{amsmath,amssymb,amsfonts}
\usepackage{algorithmic}
\usepackage{graphicx}
\usepackage{textcomp}
\usepackage{multirow}
\usepackage{graphicx}
\usepackage{gensymb}
\usepackage{hyperref}
\usepackage[ruled,vlined]{algorithm2e}
\usepackage{array} 

\IEEEoverridecommandlockouts                              

\graphicspath{ {figures/} }

\title{\LARGE \bf
Top-view Trajectories: A Pedestrian Dataset of Vehicle-Crowd Interaction from Controlled Experiments and Crowded Campus}

\author{Dongfang~Yang, Linhui Li, Keith Redmill, and \"{U}mit~\"{O}zg\"{u}ner, \IEEEmembership{Life~Fellow,~IEEE}
\thanks{Material reported here was partially supported by the United States Department of Transportation under Award Number 69A3551747111 for the Mobility21 University Transportation Center, partially by the National Natural Science Foundation of China under Grant 51775082, and partially by the China Fundamental Research Funds for the Central Universities under Grant DUT19LAB36.}
\thanks{Dongfang~Yang, Keith Redmill, and \"{U}mit~\"{O}zg\"{u}ner are with the Department of Electrical and Computer Engineering, The Ohio State University, Columbus, OH, 43210, USA. {\tt\small \{yang.3455, redmill.1, ozguner.1\}@osu.edu}}
\thanks{Linhui Li is with the School of Automotive Engineering, Faculty of Vehicle Engineering and Mechanics, Dalian University of Technology, Dalian, Liaoning, China. {\tt\small lilinhui@dlut.edu.cn}}%
}

\begin{document}

\maketitle
\thispagestyle{empty}
\pagestyle{empty}

\begin{abstract}

Predicting the collective motion of a group of pedestrians (a crowd) under the vehicle influence is essential for the development of autonomous vehicles to deal with mixed urban scenarios where interpersonal interaction and vehicle-crowd interaction (VCI) are significant. This usually requires a model that can describe individual pedestrian motion under the influence of nearby pedestrians and the vehicle. This study proposed two pedestrian trajectory datasets, CITR dataset and DUT dataset, so that the pedestrian motion models can be further calibrated and verified, especially when vehicle influence on pedestrians plays an important role. CITR dataset consists of experimentally designed fundamental VCI scenarios (front, back, and lateral VCIs) and provides unique ID for each pedestrian, which is suitable for exploring a specific aspect of VCI. DUT dataset gives two ordinary and natural VCI scenarios in crowded university campus, which can be used for more general purpose VCI exploration. The trajectories of pedestrians, as well as vehicles, were extracted by processing video frames that come from a down-facing camera mounted on a hovering drone as the recording equipment. The final trajectories of pedestrians and vehicles were refined by Kalman filters with linear point-mass model and nonlinear bicycle model, respectively, in which xy-velocity of pedestrians and longitudinal speed and orientation of vehicles were estimated. The statistics of the velocity magnitude distribution demonstrated the validity of the proposed dataset. In total, there are approximate 340 pedestrian trajectories in CITR dataset and 1793 pedestrian trajectories in DUT dataset. The dataset is available at GitHub\footnote{CITR dataset is available at  \url{https://github.com/dongfang-steven-yang/vci-dataset-citr}; DUT dataset is available at: \url{https://github.com/dongfang-steven-yang/vci-dataset-dut}}.

\end{abstract}


\section{Introduction}

In mixed urban scenarios, intelligent vehicles may have to cope with a certain number of surrounding pedestrians. In such scenarios, it is necessary to understand how vehicles and pedestrians interact with each other. This interaction has been studied for some time, but in most cases, the number of pedestrians is small so that the interpersonal interaction is usually ignored, which is not always the case in real world applications. For example, under the same vehicle influence, a group of large number of pedestrians may behave differently than a group of small number of pedestrians, because a larger group, i.e., a crowd, plays a more dominant role in the vehicle-pedestrian interaction. This vehicle-crowd interaction (VCI) scenario has been drawing attention in recent years. Specific models \cite{zeng2017specification}\cite{anvari2015modelling}\cite{yang2017agent}\cite{cheng2018mixed} have been designed to describe the individual motion of a crowd in some specific situations where both interpersonal and vehicle-pedestrian interaction were differently considered. To either calibrate or train such models above and further evaluate the their performance, providing ground truth trajectories of VCI is becoming increasingly important. However, to the best of authors' knowledge, there is no public dataset that covers VCI, especially in scenarios where interpersonal interaction is not negligible. To fill this gap, we built two VCI datasets. One (CITR dataset) focuses on fundamental VCI scenarios in controlled experiments, and the other (DUT dataset) consists of natural VCIs in crowded university campus.


In general, the approaches for modeling pedestrian motion in crowd can be classified in two categories. Traditionally, a rule-based model, e.g., social force models \cite{zanlungo2011social}, is designed based on human experience and the parameters of the model are then calibrated by using ground truth pedestrian trajectories \cite{zeng2017specification}\cite{daamen2012calibration}. Recently, with the growing popularity of deep learning, long-short term memory (LSTM) networks have been applied to model this pedestrian motion \cite{alahi2016social}\cite{pfeiffer2018data} in the hope of taking advantage of the potential in deep neural networks, which heavily relies on pedestrian trajectory data. The requirement of ground truth pedestrian trajectories in both approaches confirmed the necessity of building more pedestrian/crowd trajectory dataset, especially in scenarios that have not been covered in existing ones. Existing dataset such as ETH \cite{pellegrini2009you} and UCY \cite{lerner2007crowds} only covers interpersonal interaction, which is not suitable for VCI. Stanford Drone Dataset \cite{robicquet2016learning} includes some vehicle trajectories, but the number of surrounding pedestrians is small so that there is little interpersonal interaction. This work aims to provide a new type of pedestrian trajectory dataset that can enrich the existing datasets, and meanwhile assists in solving pedestrian safety related problems in the areas of intelligent vehicles and intelligent transportation systems.

Unlike pure interpersonal interaction, VCI introduces more complexity. This complexity can be decomposed by separating vehicle influence from interpersonal influence and by identifying different types of vehicle influence on pedestrians. To this end, in our CITR dataset, controlled experiments were designed and conducted in a way that from interpersonal interaction scenarios to VCI scenarios, they can be pairwisely compared so that separate effect, for example, the existence (or not) of a vehicle or the walking direction of the crowd, can be identified and analyzed. 

Some pedestrian motion models may consider personal characteristics, i.e., each pedestrian applies a model with a unique parameter set. CITR dataset provides such personality by assigning the same pedestrian always the same ID, hence more options are provided to researchers.

To supplement each other, in DUT dataset natural VCI data was constructed from a series of recordings of crowded university campus. A down-facing camera attached to a drone hovering above and far away from the ground was used as the recording equipment. Therefore, both the crowd and the vehicle are unaware of being observed, hence producing natural behavior. The DUT dataset can be used for final verification of VCI models or some end-to-end VCI modeling design. 

Both CITR and DUT datasets applied a hovering drone as the recording equipment. This ensured the accuracy of the extracted trajectories by avoiding the issue of occlusion, a major deficiency if pedestrians are detected from the view of sensors mounted on moving vehicles or buildings.

The trajectories of individual pedestrians and vehicles were extracted by image processing techniques. Due to the unavoidable instability of the camera attached to a hovering drone (even with a gimbal system), the recorded videos were stabilized before further processing. A robust tracking algorithm (CSRT\cite{lukevzivc2018discriminative}) was then applied to automatically track pedestrians and vehicles, although the initial positions still have to be manually selected. In the last step, different Kalman filters were applied to further refine the trajectories of both pedestrians and vehicles. This design avoided tedious manual annotation as done in the ETH and UCY dataset~\cite{pellegrini2009you}\cite{lerner2007crowds}, and possible imprecision of the tracking as done in the Stanford dataset \cite{robicquet2016learning}.

In general, the contribution of the study can be summarized as follows:
\begin{itemize}
    \item We built a new pedestrian trajectory dataset that covers both interpersonal interaction and vehicle-crowd interaction. 
    \item The dataset includes two portions. One comes from controlled experiments, in which fundamental VCIs are covered and each person has a unique ID. The other comes from crowded university campus scenarios where the pedestrian reaction to a vehicle is completely natural.
    \item The application of a drone camera for video recording, a new design of tracking strategy, and the Kalman filters for refining trajectories made the extracted trajectories as accurate as possible.
\end{itemize}

In the rest of the paper, section 2 reviews related dataset regarding pedestrian motion and vehicle-pedestrian interaction. Section 3 details the configuration of both CITR and DUT dataset. Section 4 describes the algorithm applied for trajectory extraction and the Kalman filters used for trajectory refinement. Section 5 shows some statistics of our dataset. Section 6 concludes the study and discusses possible improvement. 

\section{Related Works}

\begin{table*}
\centering
\caption{Comparison with existing world coordinate based pedestrian trajectory dataset}
\begin{tabular}{p{1cm}p{1.6cm}p{1.2cm}p{1.5cm}p{2.0cm}p{1.0cm}p{1.0cm}p{1.5cm}p{1.5cm}p{1.3cm}} 

\hline

	Dataset Name & Scenarios & Pedestrian Density & Other Participants & Method of Annotation & FPS of Annotation & Amount of Trajectories & Camera Depression Angle (degrees) & From Pixel to World Coordinate & Video Resolution \\ \hline \hline
	ETH & campus, urban street & medium & no & manual & 2.5 & 650 &  about 70-80  & matrix file & 720x576 \\ \hline
	UCY & campus, park, urban street & high, low & no & manual & interpolated & 909 & about 20-50 & partially measurable & 720x576 \\ \hline
	Stanford & campus & medium, low & cyclist, bus, golf cart, car & tracking + interpolation & 29.97 & 3297 & 90 & n.a. & 595x326 \\ \hline
	CITR & specifically designed & medium & golf cart & CSRT tracker + initial annotation & 29.97 & 340 & 90 & measured & 1920x1080 \\ \hline
	DUT & campus & high, low, medium & car & CSRT tracker + initial annotation & 23.98 & 1793 & 90 & measured & 1920x1080 \\ \hline
	Town Center & urban street & medium & no & manual + tracking verification & 25 & 2200 & about 25-35 & n.a. & 1920x1080 \\ \hline
	Train Station & train station hall & high, medium & no & KLT keypoint tracker & varied & 47866 & about 40-50 & n.a. & 720x480 \\ \hline

	
	
	

\end{tabular}
\label{tab:compare_existing_dataset}
\end{table*}

Pedestrian dataset can be in general divided into two categories: world coordinate (WC) based dataset and vehicle coordinate (VC) based dataset. WC based dataset is usually applied to studies that need to consider interpersonal interaction, because the collective motion of pedestrians is clear, accurate enough, and easily accessible, while VC based dataset doesn't contain enough instances of interpersonal interaction. Popular WC based dataset includes UCY Crowds-by-Example dataset \cite{lerner2007crowds}, ETH BIWI Walking Pedestrians dataset \cite{pellegrini2009you}, Town Center dataset \cite{benfold2011stable}, Train Station dataset \cite{zhou2012understanding} and Stanford Drone dataset \cite{robicquet2016learning}. They have been widely used for crowd motion analysis, risk detection, and the calibration/training of various rule-based and learning-based pedestrian motion models \cite{becker2018evaluation}. The proposed dataset in this study aims to enrich the WC based dataset by incorporating the vehicle-crowd interaction. A comparison among the proposed and existing WC based datasets are shown in table~\ref{tab:compare_existing_dataset}. VC based dataset is usually used for single/multiple, but not too many, pedestrian detection and/or intention estimation from a mono camera mounted in front of the vehicle. A couple of datasets such as Daimler Pedestrian Path Prediction dataset \cite{schneider2013pedestrian} and KITTI dataset \cite{Geiger2012CVPR} provide vehicle motion information, hence the trajectories of both the vehicle and pedestrians in world coordinate can be estimated by combining vehicle motion and video frames. The estimated trajectories can serve as ground truth data for vehicle-pedestrian interaction but with little interpersonal interaction due to the limited number of pedestrians. 

Some existing datasets also apply a down-facing camera attached to a hovering drone as the recording equipment. For example, in Stanford Drone dataset \cite{robicquet2016learning}, the utilization of drone eliminated occlusion so that all participants (pedestrians, cyclists, cars, carts, buses) were individually tracked. Another dataset HighD \cite{highDdataset}, which focuses on vehicle-vehicle interaction on highway driving, also successfully demonstrated the benefit of using the hovering drone to remove occlusion.

\section{Dataset}
\subsection{CITR Dataset}
\begin{figure}
    \centering
    \includegraphics[width=\linewidth]{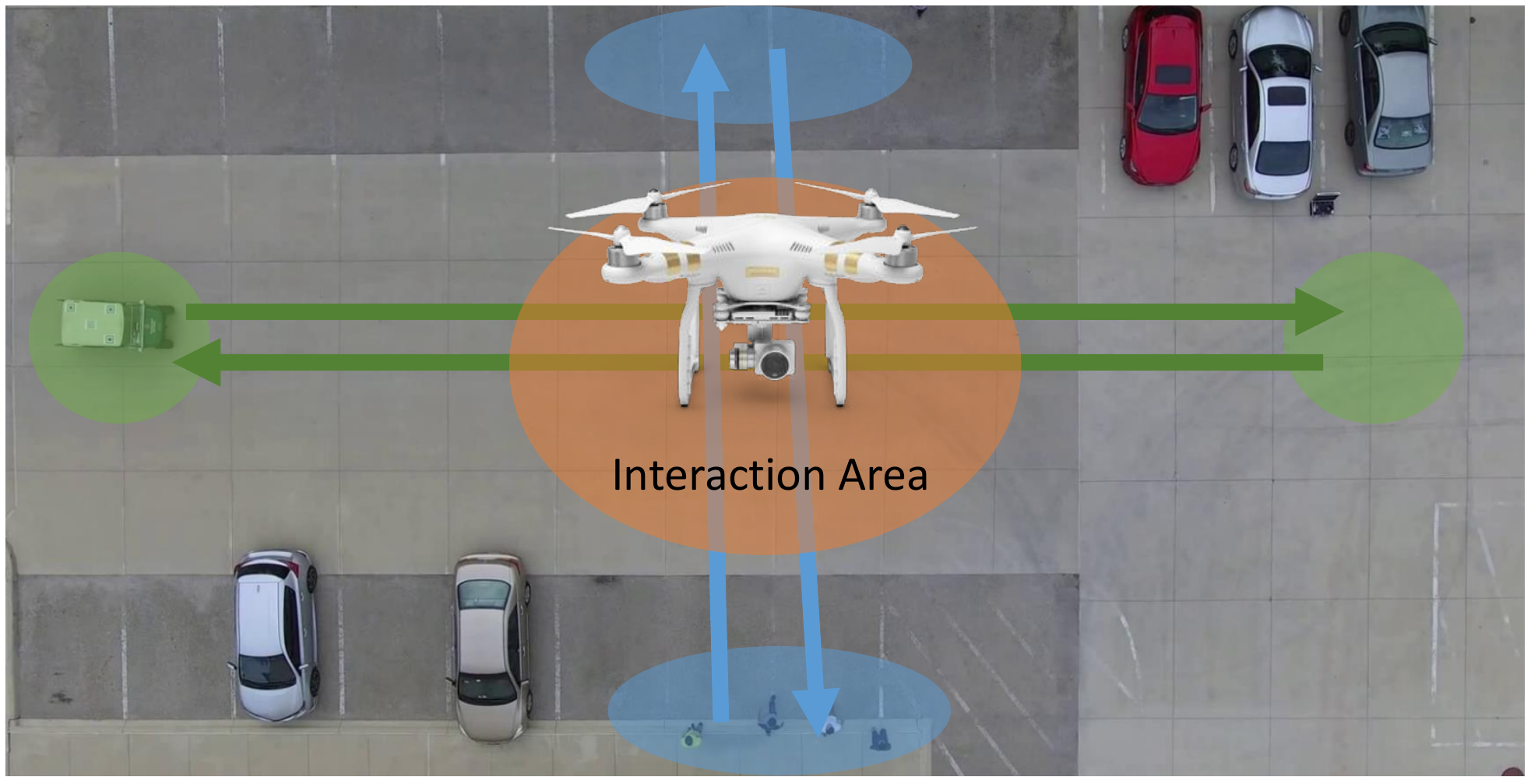}
    \caption{Layout of the controlled experiment area (a parking lot near CITR Lab at OSU). The vehicle (a golf cart) moves back and forth between two blue areas. Pedestrians move back and forth between two green areas. The interaction happens in the orange area, which is also the central area of the recording. }
    \label{fig:layout_citr}
\end{figure}
\begin{figure}
    \centering
    \includegraphics[width=\linewidth]{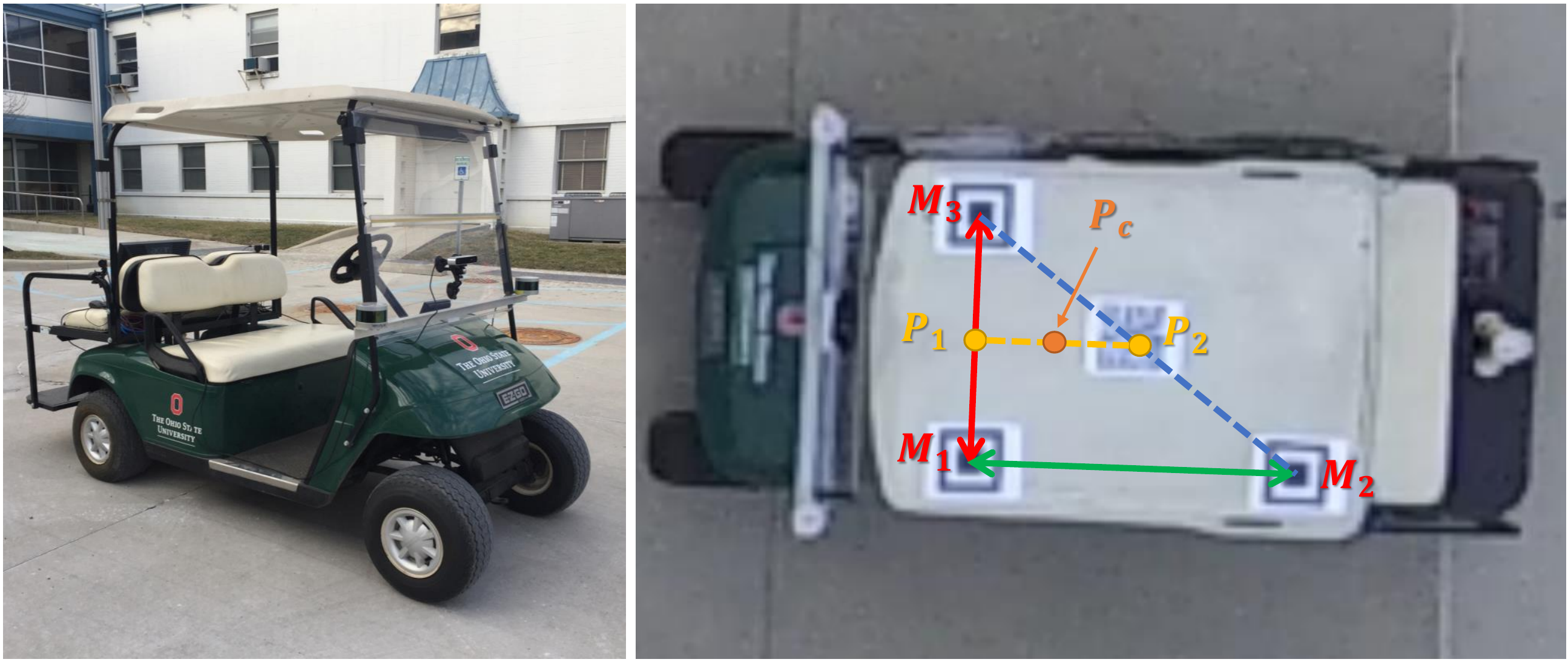}
    \caption{EZ-GO Golf cart employed in the experiments (left) and makers on top of the vehicle (right). In the vehicle tracking process, 3 markers ($M_1, M_2, M_3$) were continuously being tracked. By geometry, $P_1, P_2$ were calculated and recorded for vehicle orientation and $P_c$ as the vehicle center position.}
    \label{fig:golf_cart}
\end{figure}
The controlled experiments were conducted in a parking lot near the facility of Control and Intelligent Transportation Research (CITR) Lab at The Ohio State University (OSU). Figure~\ref{fig:layout_citr} shows the layout of the experiment area. A DJI Phamton 3 SE Drone with a down-facing camera on a gimbal system was used as the recording equipment. The video resolution is $1920\times1080$ with an fps of 29.97. Participants are the members of CITR Lab at OSU. During the experiments, they were instructed only to walk from one small area (starting points) to another small area (destinations). The employed vehicle was an EZ-GO Golf Cart, as shown in figure~\ref{fig:golf_cart}. 3 markers were put on top of the vehicle to help vehicle motion tracking, of which the vehicle position is calculated by geometry. The reason of using 3 markers is to reduce the tracking noise as much as possible. 
\begin{figure}
    \centering
    \includegraphics[width=\linewidth]{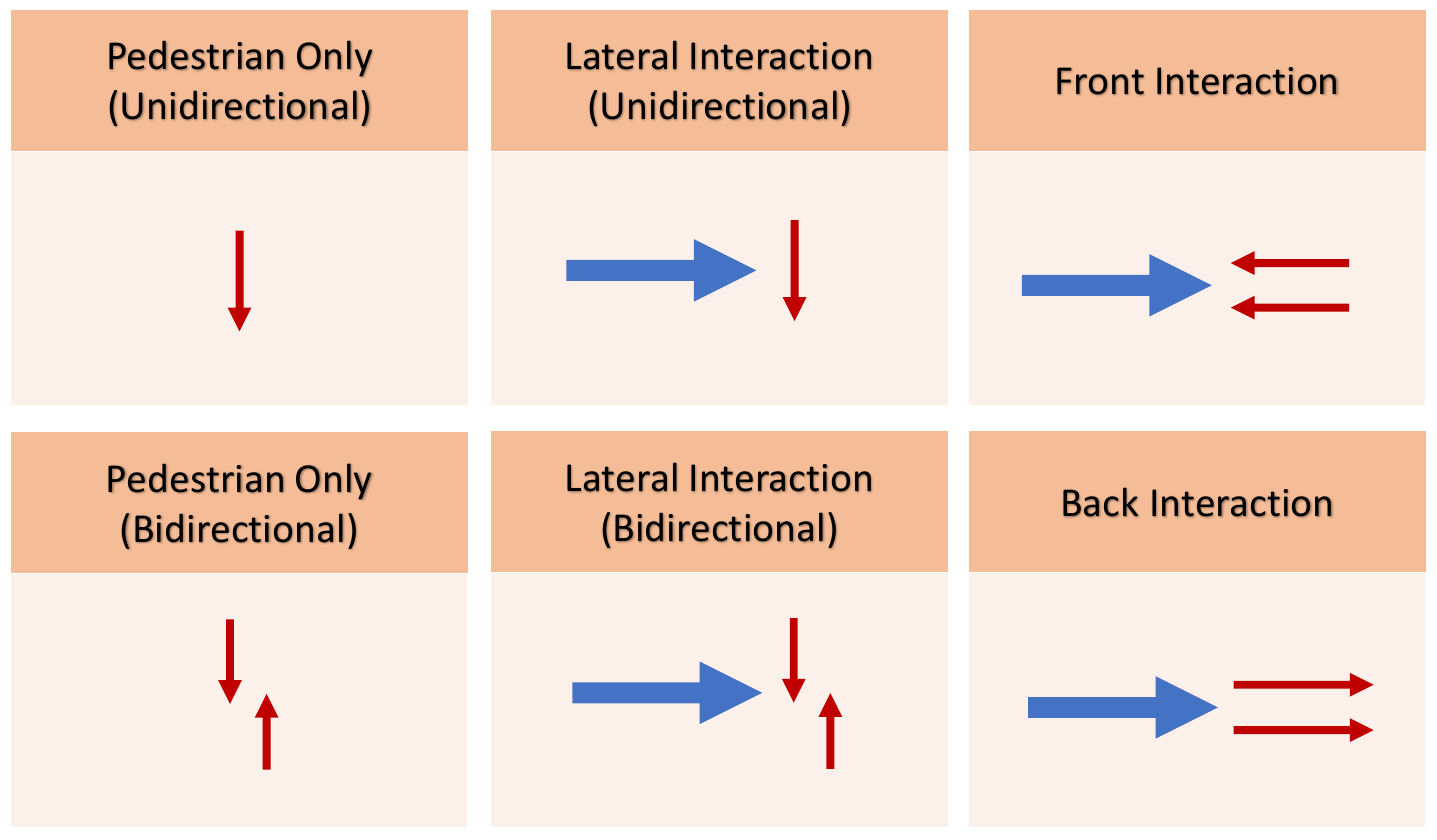}
    \caption{Designed scenarios of controlled experiments. Red arrows indicate the motion of pedestrians/crowd, while blue arrows indicate vehicle motion.}
    \label{fig:scenarios_citr}
\end{figure}

The designed fundamental scenarios were generally divided into 6 groups, as shown in figure~\ref{fig:scenarios_citr}. They were designed such that by comparing pedestrian-only scenarios (pure interpersonal interaction) and VCI scenarios, the vehicle influence can be separated and analyzed. Therefore, except for the difference due to the existence (or not) of a vehicle, all other factors remain the same such as pedestrians' intention (starting point and destination), pedestrians' identity (who are these pedestrians), and environment layout (location, time period, weather, etc.). The scenarios of front, back, and side interactions help exploring typical VCIs which could guide the design of pedestrian motion models.

After processing, there are 38 video clips in total, which include approximate 340 pedestrian trajectories. The detailed information is presented in table~\ref{tab:scenarios_citr}.

\begin{table}
\centering
\caption{Number of clips in each scenario of CITR dataset}
\begin{tabular}{cc} 
    \hline
	Scenarios & Num. of clips \\
	\hline\hline
	Pedestrian only (unidirectional) & 4 \\ \hline
	Pedestrian only (bidirectional) & 8 \\ \hline
	Lateral interaction (unidirectional) & 8 \\ \hline
	Lateral interaction (Bidirectional) & 10 \\ \hline
	Front interaction & 4 \\ \hline
	Back interaction & 4 \\ \hline
\end{tabular}
\label{tab:scenarios_citr}
\end{table}

\subsection{DUT Dataset}
\begin{figure}
    \centering
    \includegraphics[width=\linewidth]{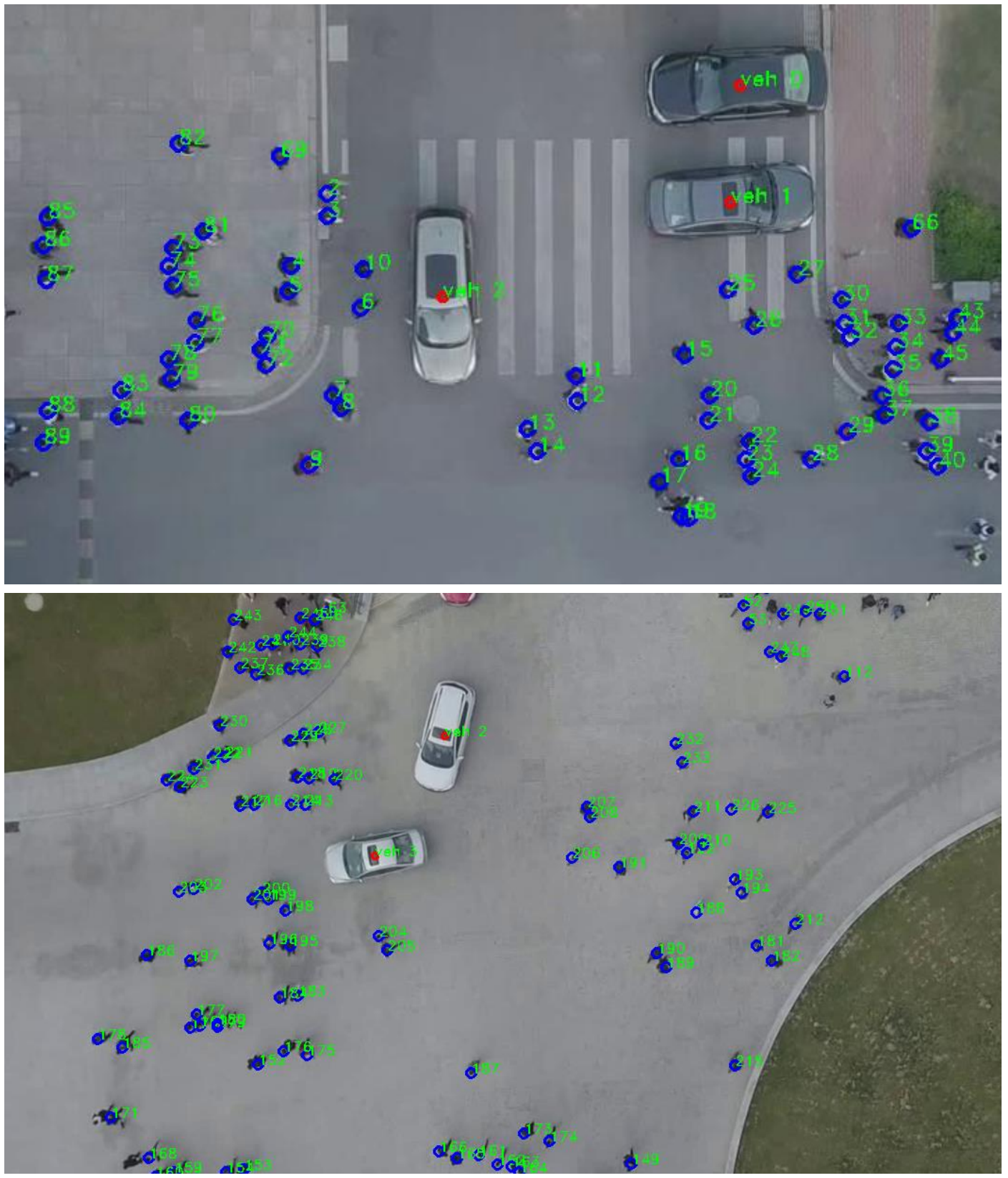}
    \caption{Locations of DUT dataset. Upper: an area of crosswalk at an intersection without traffic signals. Lower: a shared space near a roundabout.}
    \label{fig:scenarios_dut}
\end{figure}

The DUT dataset was collected at two crowded locations in the campus of Dalian University of Technology (DUT) in China, as shown in figure~\ref{fig:scenarios_dut}. One location includes an area of pedestrian crosswalk at an intersection without traffic signals. When VCI happens, in general there is no priority for either pedestrians or vehicles. The other location is a relatively large shared space near a roundabout, in which pedestrians and vehicles can freely move. Similar to CITR dataset, a DJI Mavic Pro Drone with a down-facing camera was hovering above the interested area as the recording equipment, high enough to be unnoticed by pedestrians and vehicles. The video resolution is $1920\times1080$ with an fps of 23.98. Pedestrians are primarily made up of college students who just finished classes and on their way out of classrooms. Vehicles are regular cars that go through the campus. 

With this configuration, scenarios of DUT dataset consists of natural VCIs, in which the number of pedestrians varies hence introducing some variety of the VCI. 

After processing, there are 17 clips of crosswalk scenarios and 11 clips of shared space scenarios, including 1793 trajectories. Some of the clips contains multiple VCIs, i.e., more than 2 vehicles interacting with pedestrians simultaneously, as in the lower picture in figure~\ref{fig:scenarios_dut}.

\begin{figure}
    \centering
    \includegraphics[width=0.8\linewidth]{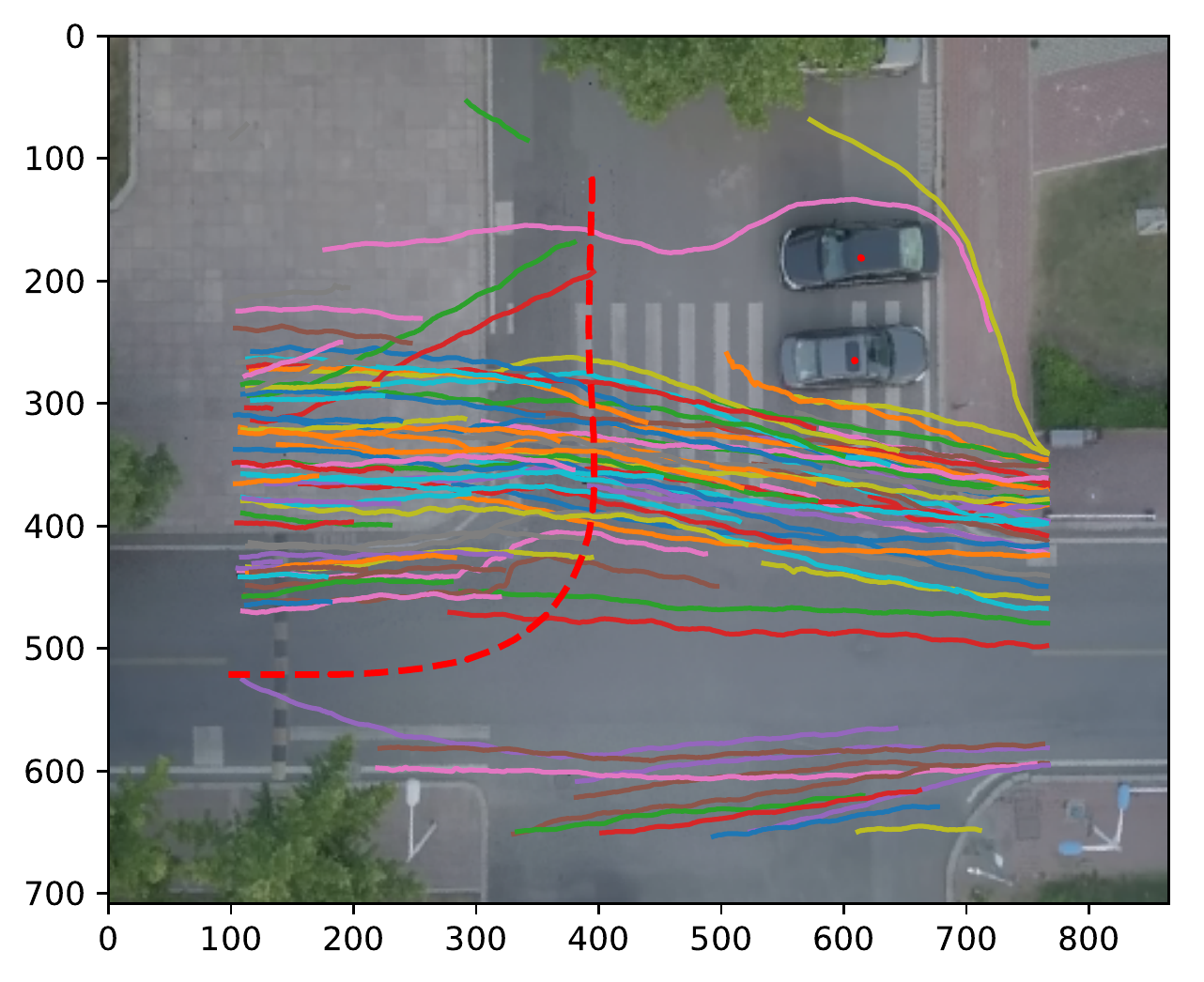}
    \caption{Trajectories of vehicles (red dashed line) and pedestrians (colorful solid lines) in a clip of the intersection scenario.}
    \label{fig:traj_demo_intersection}
\end{figure}

\begin{figure}
    \centering
    \includegraphics[width=0.8\linewidth]{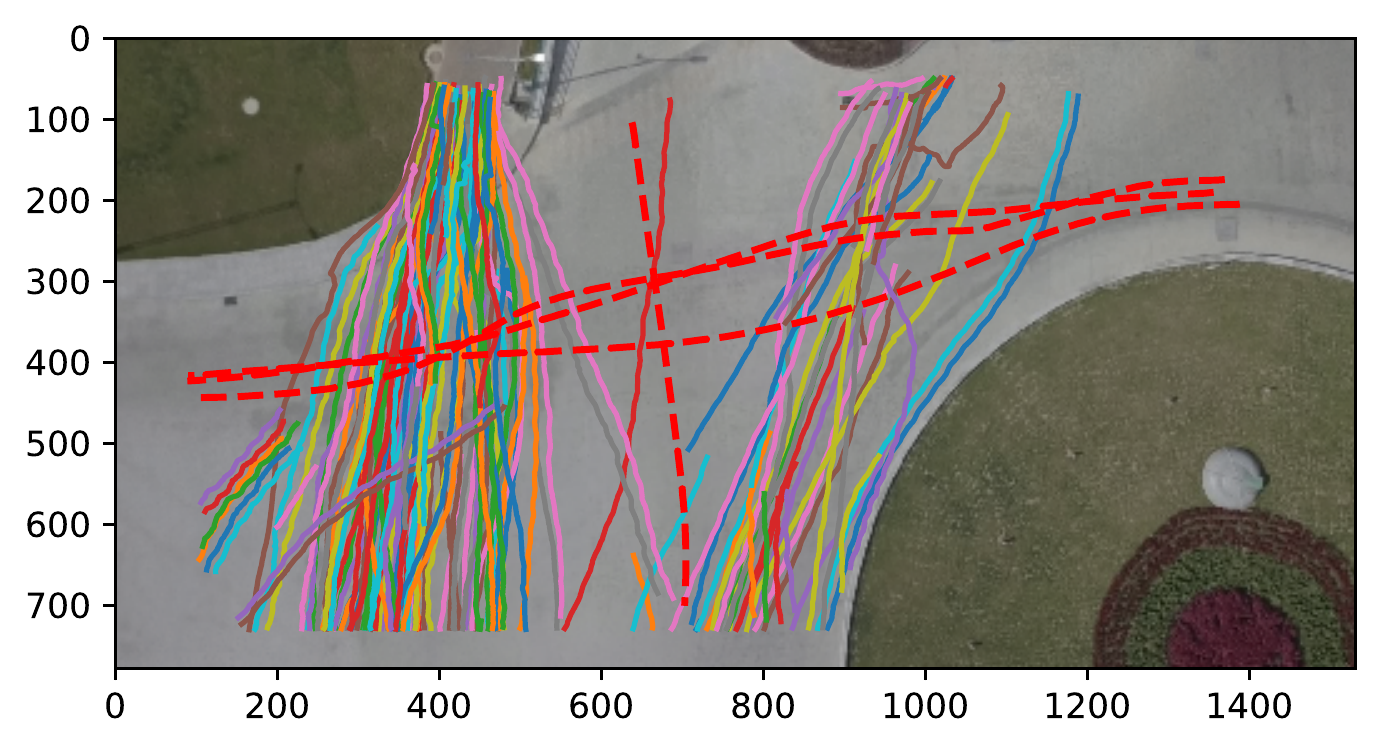}
    \caption{Trajectories of vehicles (red dashed line) and pedestrians (colorful solid lines) in a clip of the shared space scenario.}
    \label{fig:traj_demo_roundabout}
\end{figure}

Figure \ref{fig:traj_demo_intersection} and \ref{fig:traj_demo_roundabout} demonstrate the processed example trajectories of the DUT dataset. 

\section{Trajectory Extraction}
Four procedures were done to extract the trajectories of both pedestrians and vehicles from the recorded top-view video.
\begin{algorithm}
    \label{al:calibration}
    \SetAlgoLined
    \caption{Video Stabilization}
    \KwResult{calibrated frames $F_i^{cal}$}
        set 1st frame $F_1$ as reference $F_{ref}$\;
        \For{each new frame $F_i,i=2,3,\cdots$}{
            apply SIFT to find key-points in $F_i$ and $F_{ref}$, separately\;
            apply KNN to find matches\;
            obtain good matches by removing matches that have long distance of pixel positions in $F_i$ and $F_{ref}$\;
            apply RANSAC for the good matches to calculate the transformation matrix $M_i$ from $F_i$ to $F_{ref}$\;
            obtain $F_i^{cal}$ by applying transformation $M_i$ to $F_i$\;
        }
\end{algorithm}

\subsection{Video Stabilization}
First, the raw video was stabilized to remove the noise caused by unstable drone motion. This procedure applies several image processing techniques, which include scale-invariant feature transform (SIFT) algorithm for finding key-points, k-nearest neighbors (k-NN) for obtaining matches, and random sample consensus (RANSAC) for calculating perspective transformation between each video frame and the first video frame (reference frame). The detailed procedure is illustrated in algorithm~\ref{al:calibration}. 

\subsection{Vehicle and Pedestrian Tracking}
Once the video was stabilized, pedestrians and vehicles were automatically tracked by using Discriminative Correlation Filter with Channel and Spatial Reliability (CSR-DCF)~\cite{lukevzivc2018discriminative}. In the tracking process, raw videos are partitioned into small clips, which contain separate and complete VCIs. Once pedestrians appear in the region of interest (ROI), the initial positions were manually given, hence initializing the trackers. When they exited the ROI, the trackers stopped. Due to the vehicle size, vehicle tracking was done by individually tracking either the 3 markers on top of the vehicle (CITR dataset) or four corners of vehicle (DUT dataset). Then, the vehicle position was calculated based on geometric relationship of these tracked points.

\subsection{Coordinate Transformation}
Pedestrian trajectories obtained in the previous step are in the coordinates of image pixels. A coordinate transformation operation is necessary to convert the trajectories from image pixels into actual scale in meters. 

This can be done by either measuring the actual length of a relatively long reference line in the scene or measuring the distance between markers on top of the vehicle (if applicable). The assumption here is that, compared with the altitude of the hovering drone, the distance between the ground plane and the tracking plane (the plane of a pedestrian's head or the vehicle's top) is very small so that both planes can be treated as the same plane.  

\subsection{Trajectory Filtering}
In the last step, Kalman filters~\cite{faragher2012understanding} was applied to remove the noise and refine the trajectories. It is sufficient to use a linear Kalman filter with a point-mass model for pedestrian trajectories, in which the 2D velocity (in x and y axes) can be estimated. The state transition and measurement follows the equations:
\begin{align}
    \dot{x}&=v+w_1 \\
    \dot{v}&=a+w_2 \\
    y&=x + v,
\end{align}
where position $x\in \mathbb{R}^2$ and velocity $v\in \mathbb{R}^2$ are the system state, $y\in \mathbb{R}^2$ is the measurement (recorded position), $w=[w_1^T,w_2^T]^T\sim N(0,Q)$ the state transition noise, and $v\sim N(0,R)$ the measurement noise.

When applying the Kalman filter, it is assumed that $a=0$, which implies a constant velocity model.


Vehicle motion is somehow constrained, e.g., the lateral motion/velocity can not be abruptly changed. Therefore, an extended Kalman filter with a nonlinear kinematic bicycle model was applied. The bicycle model follows:
\begin{align}
\dot{x_x}&=v\cos (\theta+\beta)+w_1\\
\dot{x_y}&=v\sin (\theta+\beta)+w_2\\
\dot{\theta}&=\frac{v}{l_r}\sin \beta+w_3\\
\dot{v}&=a+w_4\\
\beta&=\arctan\left(\frac{l_r}{l_f+l_r}\tan\delta_f\right)\\
y&=[x_x,x_y]^T+v,
\end{align}
where $x_x, x_y$ stands for the position, $v$ is the longitudinal speed, $\beta$ is the velocity angle with respect to the vehicle C.G., $l_f$, $l_r$ are the distances from C.G. to the front wheel and the rear wheel, respectively, $a$ is the longitudinal acceleration, $\delta_f$ is the steering angle of the front wheel, $w=[w_1,w_2,w_3,w_4]^T\sim N(0,Q)$ the state transition error, and $v\sim N(0,R)$ the measurement error.

At each step of the extended Kalman filter, the system is linearized at current state by calculating its Jacobian. It is assumed that both inputs $a=0$ and $\delta_f=0$.

\section{Statistics}

To give a more detailed description of the above dataset, the magnitude of pedestrian velocities (estimated by the Kalman filter) in all video clips were analyzed. The reason of analyzing velocity magnitude is that, pedestrian velocity is the most intuitive way of describing pedestrian motion, and, as argued in~\cite{becker2018evaluation}, if pedestrian trajectories are used to train neural network based pedestrian model, using pedestrian velocity (offset in motion at the next time step) is better than using absolute position, because different reference systems (how the global coordinates are defined) in different dataset usually cause incompleteness of training data.

\begin{figure}
    \centering
    \includegraphics[width=\linewidth]{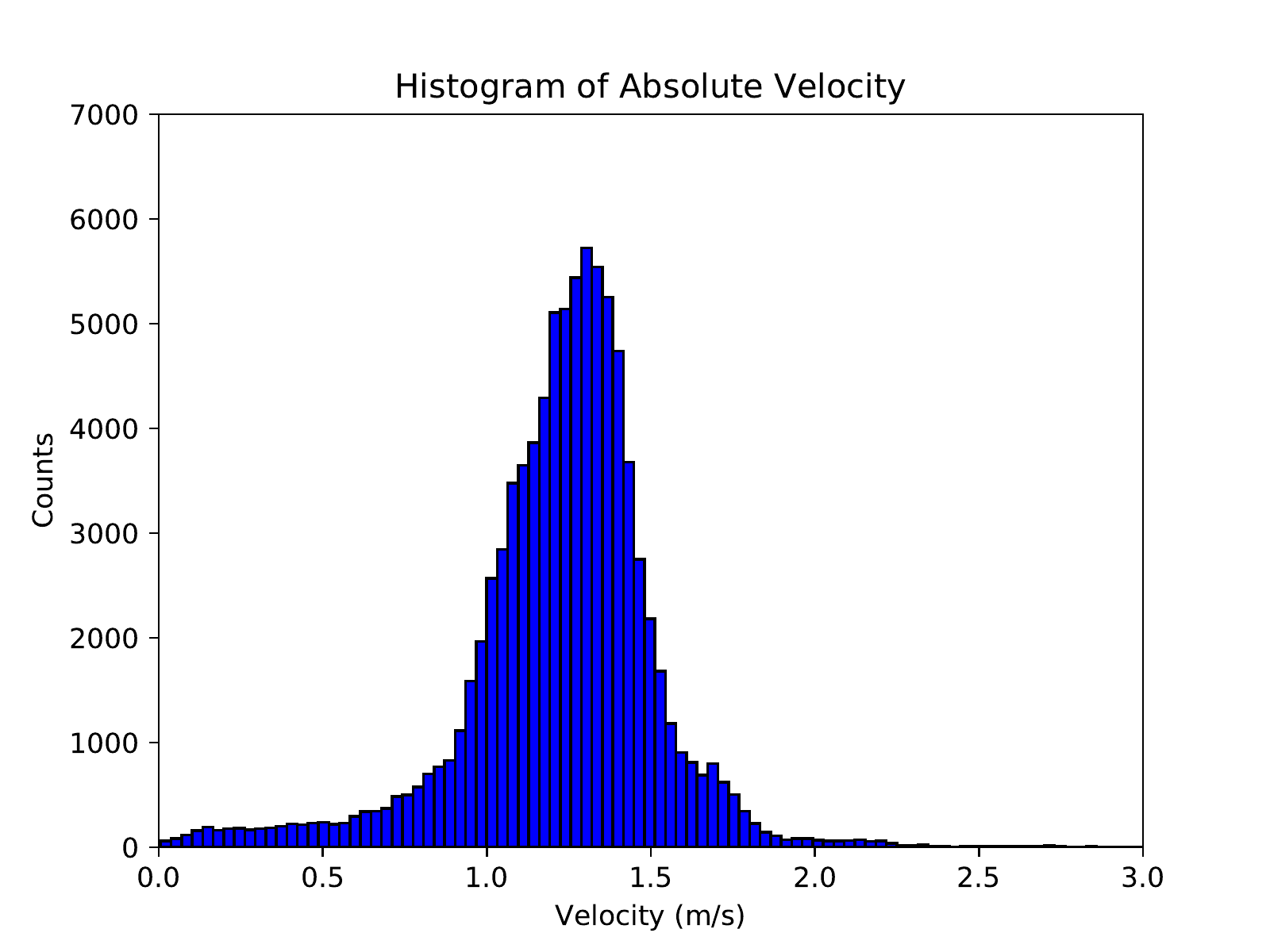}
    \caption{Distribution of velocity magnitude in CITR dataset}
    \label{fig:velocity_dist_citr}
\end{figure}

\begin{figure}
    \centering
    \includegraphics[width=\linewidth]{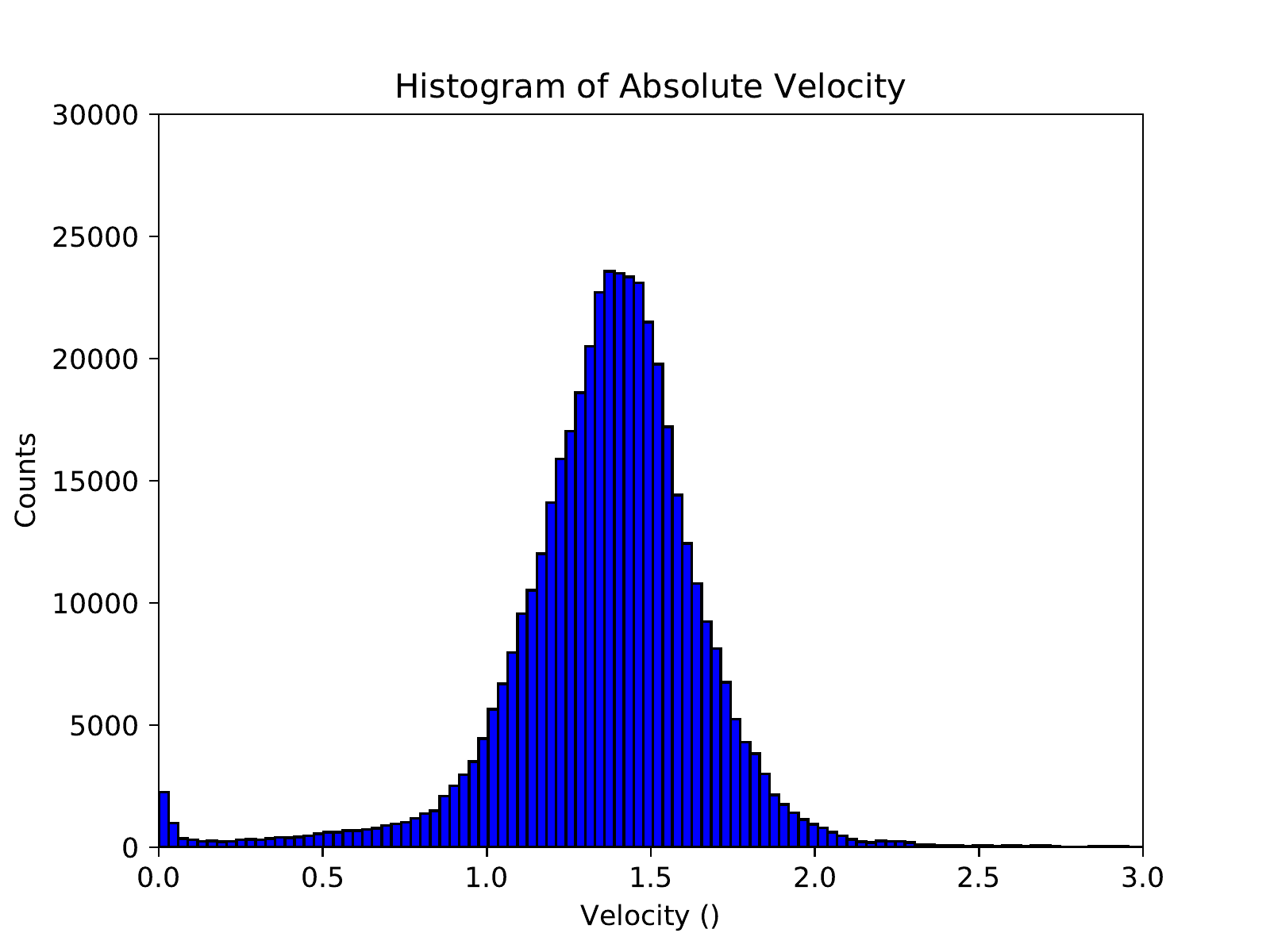}
    \caption{Distribution of velocity magnitude in DUT dataset}
    \label{fig:velocity_dist_dut}
\end{figure}

Figure \ref{fig:velocity_dist_citr} and \ref{fig:velocity_dist_dut} show the distribution of the velocity magnitude for CITR dataset and DUT dataset, respectively. Table \ref{tab:mean_velocities} presents the mean velocity magnitude and mean walking velocity magnitude. The walking velocity excludes the velocity magnitude that is less than $0.3m/s$, at which the pedestrian is considered as either standing or yielding to the vehicle instead of walking. The value of $0.3m/s$ was intuitively selected based on the shape of the histogram. It is obvious that, from the velocity distribution and the mean velocity results, the pedestrians in DUT dataset walk faster than the pedestrians in CITR dataset. The reason could be that, when conducting controlled experiments, as in the CITR dataset, pedestrians were more relaxed, while in the DUT dataset, pedestrians were in a little bit hurry because they just came out of classes. However, in general, the distribution and the mean velocity magnitude are in accordance with the preferred walking velocity in various situations \cite{mohler2007visual}.


\begin{table}
\centering
\caption{Mean velocity magnitude}
\begin{tabular}{ccc} 
    \hline
	Dataset & Mean velocity & Mean walking velocity \\
	\hline\hline
	CITR & 1.2272 & 1.2435 \\ \hline
	DUT & 1.3661 & 1.3825 \\ \hline

\end{tabular}
\label{tab:mean_velocities}
\end{table}

\section{Conclusion}

Two dataset, experimentally designed CITR dataset and natural DUT dataset, were built in this study for pedestrian motion models that consider both interpersonal and vehicle-crowd interaction. The trajectories of pedestrians and vehicles were extracted by image processing techniques and refined by Kalman Filters. The statistics of the velocity magnitude validated the proposed dataset. 

This study can be regarded as an initial attempt to incorporate VCI into pedestrian trajectory dataset. The amount of the trajectories and the variety of VCI scenarios are somehow limited, therefore, it is expected to build more dataset of various scenarios. It is also expected to build a benchmark that tests a couple of famous pedestrian motion models, which is our major future work. Another improvement could be automatically detecting/selecting initial positions of pedestrians when they entered the ROI, hence totally removing manual operation. From the aspect of personal characteristics, it would help if the pedestrians in the dataset could be identified according to their age, gender, head direction, and other features, although manual annotation of these features seems to be the only option at current stage. 

\addtolength{\textheight}{-12cm}   




\section*{ACKNOWLEDGMENT}

The authors would like to thank Xinran Wang for collecting and processing DUT dataset, and Ekim Yurtsever and John Maroli for conducting controlled experiments. Also many thanks to the members who have supported the dataset building at Control and Intelligent Transportation Research (CITR) Lab at The Ohio State University (OSU) and at Dalian University of Technology (DUT).


\bibliographystyle{ieeetr}
\bibliography{mybib}

\end{document}